\documentclass[sigconf]{acmart}

\usepackage{makecell}                
\usepackage{booktabs}  

\usepackage{multirow}                 
\usepackage{multicol}                 
\usepackage{multirow} 
\AtBeginDocument{%
  \providecommand\BibTeX{{%
    \normalfont B\kern-0.5em{\scshape i\kern-0.25em b}\kern-0.8em\TeX}}}

\setcopyright{none}
\renewcommand\footnotetextcopyrightpermission[1]{}
\copyrightyear{2023}
\acmYear{2023}
\setcopyright{acmlicensed}\acmConference[MMAsia '23]{ACM Multimedia Asia
2023}{December 6--8, 2023}{Tainan, Taiwan}
\acmBooktitle{ACM Multimedia Asia 2023 (MMAsia '23), December 6--8, 2023,
Tainan, Taiwan}
\acmPrice{15.00}
\acmDOI{10.1145/3595916.3626441}
\acmISBN{979-8-4007-0205-1/23/12}
\acmConference{ACM Multimedia Asia (MMAsia'23)}{December 06--08,
  2023}{Tainan, Taiwan}
\acmPrice{15.00}
\acmISBN{978-1-4503-XXXX-X/18/06}

\begin{document}

\title{RGB-D Tracking via Hierarchical Modality Aggregation and Distribution Network}



\author{Boyue Xu}
\affiliation{%
  \institution{State Key Laboratory for Novel Software Technology, Nanjing University}
  \city{Nanjing}
  \country{China}
}
\email{xuby@smail.nju.edu.cn}

\author{Yi Xu}
\affiliation{%
  \institution{State Key Laboratory for Novel Software Technology, Nanjing University}
  \city{Nanjing}
  \country{China}
}
\email{yxu1025@smail.nju.edu.cn}

\author{Ruichao Hou}
\affiliation{%
  \institution{State Key Laboratory for Novel Software Technology, Nanjing University}
  \city{Nanjing}
  \country{China}
}
\email{rc\_hou@smail.nju.edu.cn}

\author{Jia Bei}
\authornote{Corresponding author.}
\affiliation{%
  \institution{State Key Laboratory for Novel Software Technology, Nanjing University}
  \city{Nanjing}
  \country{China}
}
\email{beijia@nju.edu.cn}

\author{Tongwei Ren}
\affiliation{%
  \institution{State Key Laboratory for Novel Software Technology, Nanjing University}
  \city{Nanjing}
  \country{China}
}
\email{rentw@nju.edu.cn}

\author{Gangshan Wu}
\affiliation{%
  \institution{State Key Laboratory for Novel Software Technology, Nanjing University}
  \city{Nanjing}
  \country{China}
}
\email{gswu@nju.edu.cn}






\renewcommand{\shortauthors}{Xu \emph{et al.}}

\begin{abstract}
The integration of dual-modal features has been pivotal in advancing RGB-Depth (RGB-D) tracking. However, current trackers are less efficient and focus solely on single-level features, resulting in weaker robustness in fusion and slower speeds that fail to meet the demands of real-world applications. In this paper, we introduce a novel network, denoted as HMAD (Hierarchical Modality Aggregation and Distribution), which addresses these challenges. HMAD leverages the distinct feature representation strengths of RGB and depth modalities, giving prominence to a hierarchical approach for feature distribution and fusion, thereby enhancing the robustness of RGB-D tracking. Experimental results on various RGB-D datasets demonstrate that HMAD achieves state-of-the-art performance. Moreover, real-world experiments further validate HMAD's capacity to effectively handle a spectrum of tracking challenges in real-time scenarios.
\end{abstract}





\maketitle

\section{Introduction}
 RGB-D tracking is a type of multimodal object tracking~\cite{hou1,hou2,radar2,radar3}, which combines RGB and depth data. RGB offers visual details like color and texture but is limited to 2D dimensions. Depth information, providing the distance from the camera to the object, enables accurate 3D object localization. This combination is widely used in tasks like saliency detection, object detection, and tracking~\cite{review, ren2016, Ren2019, rgbddet, rgbd1k}.

Among these tasks, RGB-D tracking presents the most valuable and challenging applications. It plays a crucial role in human-computer interaction, robotic environmental perception, and other domains~\cite{human,rgbdact,app1,app2}. The fundamental requirement of this task is to efficiently utilize the complementary dual-modal features in real-time to address complex tracking scenarios. Especially in complex indoor environments with intricate backgrounds, similar targets, and dim lighting, the optimal use of these dual-modal complementary features becomes the central challenge for RGB-D tracking.

Current RGB-D trackers primarily focus on modal fusion methods~~\cite{DepthTrack,rgbd1k,protrack}. Fusion methods have evolved from simple additive or weighted fusion of both modalities~\cite{DepthTrack} to using Transformer for fusion~\cite{rgbd1k} or applying prompt learning to guide the fusion process~\cite{VIPT}. While these methods have gradually improved fusion quality, they have introduced two challenges: (1) Can the increasingly complex fusion methods meet the real-time requirements of RGB-D tracking tasks? This concern is particularly relevant in contexts where RGB-D tracking is commonly used, such as robotics and human-computer interaction device, where computational power is limited. (2) Are all features suitable for RGB-D tracking tasks? Is it possible to select appropriate features for the tracking task? 

\begin{figure}[t]
    \centering
    \includegraphics[scale=0.45]{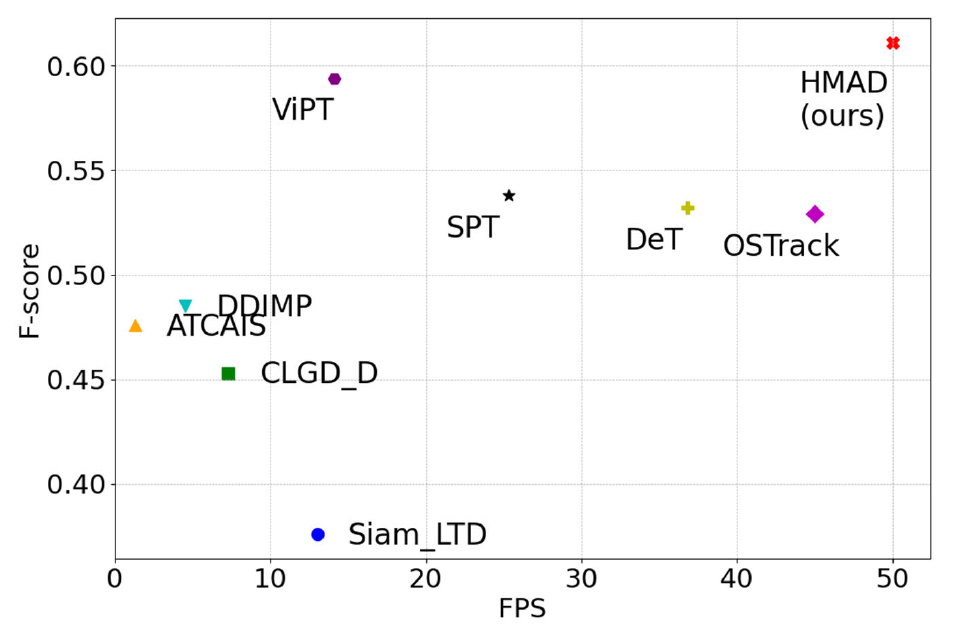}
    \caption{Comparison results with representative trackers on DepthTrack dataset.}
    \label{fig:comp}
\end{figure} 
We propose the HMAD tracker specifically designed to address the aforementioned issues. To address the first issue, the HMAD tracker adopts the classical RGB tracking framework DIMP~\cite{DIMP} as a baseline. This framework has strong feature discrimination capability and uses less computational cost compared to the complex Transformer structure. Additionally, HDMA itself is a simple and effective hierarchical modality fusion method, avoiding the substantial speed loss that complex fusion methods can entail. For the second problem, we propose a hierarchical modality aggregation and distribution network to extract rich color and texture features provided by RGB at the shallow feature level, and semantic features such as contour distance provided by depth information at the deep feature level. By effectively integrating these features, we extract as many valid features as possible while using minimal resources, thus enhancing the robustness of RGB-D tracking. Experimental results show that the HMAD tracker performs well on major RGB-D tracking datasets. Importantly, it achieves a tracking speed of 15 FPS on edge device, meeting real-time tracking requirements.

The main contributions of this paper is summarized as follows:
\begin{itemize}
    \item  We propose a real-time RGB-D tracker HMAD based on the DIMP framework, which guarantees high tracking performance while achieving real-time tracking speed.

    \item  We propose a hierarchical modality aggregation and
distribution network capable of fully extracting and integrating effective features from both RGB and depth information.

    \item  We validate the effectiveness of the HMAD on existing RGB-D tracking datasets. Moreover, we conduct real-world experiments on real-world edge device to demonstrate the effectiveness and real-time capabilities of the tracker in real-world scenarios.
\end{itemize}
\section{RELATED WORK}
\subsection{Single Object Tracking}
Two primary strategies dominate the field of single-object tracking: Siamese-based and tracking-by-detection methods.  Siamese-based methods, represented by SiamFC~\cite{fc}, utilize template correlation within a search area to identify tracking targets. Subsequent enhancements to this approach include SiamRPN~\cite{rpn} and SiamRPN++~\cite{rpn++} by Li \emph{et al.}, which integrate RPN~\cite{frcnn} networks and feature pyramids to increase model performance. Further advances have seen the incorporation of Transformer~\cite{Attention} structures, as demonstrated in Chen \emph{et al.}'s TransT~\cite{TransT} tracking network, providing improved global association capabilities for tracking. Subsequently, Cui \emph{et al.} delved into the Transformer architecture, going a step further to replace the traditional convolutional neural network backbone with a Transformer structure~\cite{Attention}, which led to the development of the MixFormer~\cite{MixFormer}. This innovation greatly enhanced the effectiveness of feature extraction and achieved state-of-the-art results in the domain of generic single-object tracking. In a parallel approach, Lin \emph{et al.} adopted a similar line of thinking, designing a new Transformer-based backbone network for tracking, as described in their work SwinTrack~\cite{swintrack}.

Contrastingly, tracking-by-detection methods, exemplified by MDNet~\cite{MDNet} from Nam \emph{et al.}, employ a discriminator model to locate tracking target within input image. Extensions of this approach include the ATOM~\cite{ATOM} by Danelljan \emph{et al.}, which integrates both target estimation and online update modules to increase robustness. Bhat \emph{et al.}'s DiMP~\cite{DIMP} method introduced a discriminative learning loss, enhancing target identification. Moreover, the Keep-track~\cite{kp} method by Mayer \emph{et al.} incorporates a target association module that is specialized for scenarios with a large number of similar interfering targets.

\subsection{RGB-D Tracking}
In the domain of RGB-D tracking, there are two main strategies: methods based on correlation filters, and methods relying on transfer RGB tracking~\cite{review}.

The former approach includes the work of researchers such as Camplani \emph{et al.}, who proposed a real-time tracker~\cite{Real} based on the Kernelized Correlation Filters (KCF)~\cite{kcf} algorithm, which integrates RGB and depth information to manage complex situations such as occlusions. This work demonstrates promising results. Kart \emph{et al.} seek to transform short-term RGB trackers into RGB and depth trackers by proposing a general framework~\cite{How} that integrates an occlusion detection module based on depth segmentation into a correlation filter framework~\cite{Depth18}. In addition, Harika \emph{et al.} introduced multimodal fusion at different layers to integrate the features of RGB images and depth maps~\cite{hier}.

On the other hand, among the transfer RGB tracker based methods, Yan \emph{et al.} performed fusion of RGB and depth features on end-to-end RGB tracker ATOM~\cite{ATOM} and DiMP~\cite{DIMP}. They used a straightforward addition method to merge the deep features of RGB and depth images~\cite{DepthTrack}. Xue \emph{et al.} went a step further and used the more advanced Transformer tracking framework for feature fusion~\cite{rgbd1k}. They leverage the powerful feature association capability of Transformer to achieve better results, although the complexity of the Transformer structure results in slower tracking. Additionally, some researchers applied prompt learning to RGB-D tracking~\cite{protrack,VIPT}. By utilizing minimal prompt information, they were able to transfer single-modality tracking methods to RGB-D tracking, substantially reducing the complexity and challenges of training. This approach represents a novel utilization of existing methodologies, expanding the applicability to more advanced and diverse tracking scenarios.

\begin{figure*}[t]
    \centering
    \includegraphics[scale=0.6]{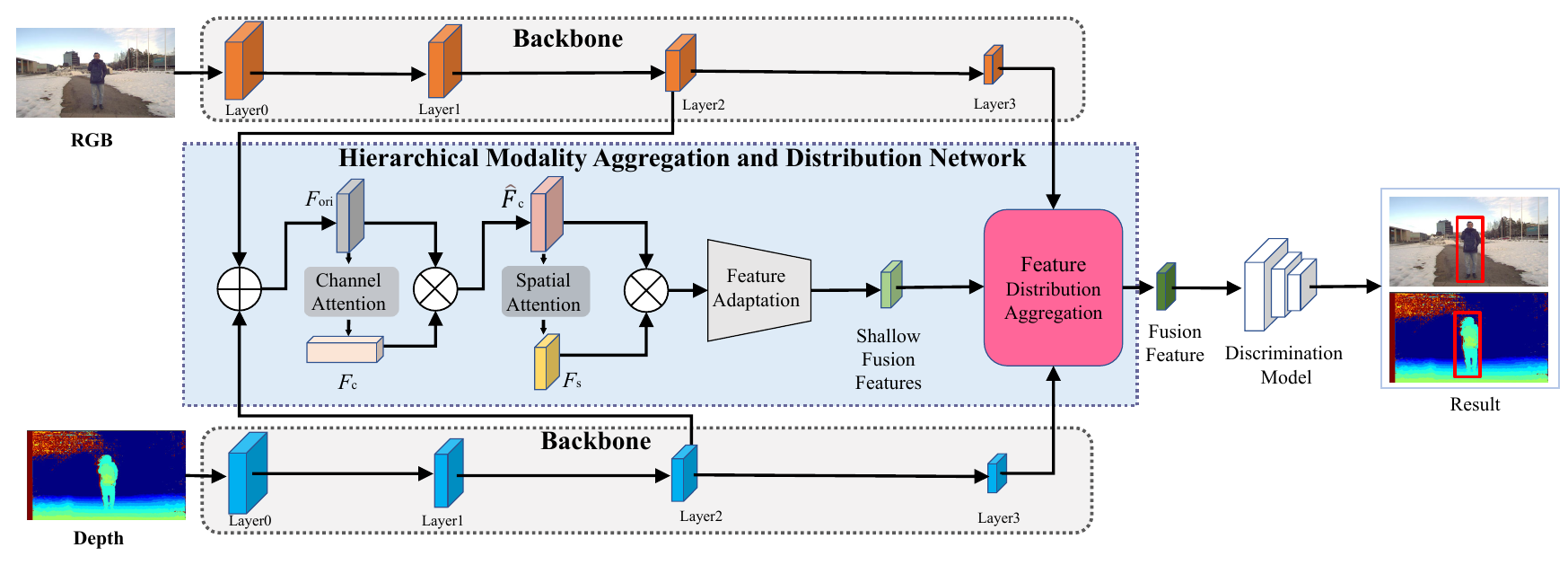}
    \caption{The framework of HMAD, consists of backbone , hierarchical modality aggregation and distribution network and a target discrimination model.}
    \label{fig:pipe}
\end{figure*} 
\section{METHODOLOGY}


\subsection{Network Architecture}
We propose an RGB-D single object tracking method namde HMAD, based on the DIMP~\cite{DIMP}. This method incorporates a hierarchical modality aggregation and
distribution network to integrate features from both modalities. These fused features are then used to train a discrimination model, which subsequently predicts the classification score and location of the tracking target. The flow of the method is illustrated in Figure ~\ref{fig:pipe}.

The input of the HMAD tracker is the corresponding RGB and depth images, which are merged through a hierarchical modality aggregation and distribution network. This process consists of two parts: the first part involves attention-based~\cite{cbam} shallow feature extraction, while the second part involves feature distribution and fusion. The former is responsible for extracting the effective components in shallow features, while the latter is responsible for distributing and fusing shallow features with deep features. The fused features are then used for final target discrimination and tracking.

Our method uses random samples from video sequences to train a discriminative model derived from the DIMP~\cite{DIMP}. The fused features are fed into the discrimination model initialization module, which initializes the target region. This module adopts a method of precise pooling to generate \emph{W} $\times$ \emph{H} $\times$ \emph{N} feature filters. It then iteratively optimizes the initialized filters with background information from the target region and generates the final discriminative model, which is used in the testing phase to predict the classification score and location of the tracked target.

In order to maintain the robust generalization ability of the discrimination model during tracking, the HMAD tracker employs the online update strategy of DIMP~\cite{DIMP} to update the discrimination model. During prediction, for the first frame with annotations, a data augmentation strategy is used to construct an initial set of \emph{K} samples, given the accuracy of the annotations. The trained discrimination model is then applied to these \emph{K} samples for further gradient descent, initially satisfying the feature distribution of the sequence. As tracking progresses, if the confidence in the tracked target is high, the network opts to discard the oldest samples, maintaining a maximum of \emph{L} samples. Throughout the tracking process, the discrimination model is optimized using the saved samples every \emph{P} frames. This ensures that the discrimination model always satisfies the feature distribution of the sequence and does not lose its discriminative power over the target features due to environmental disturbances or target scale variations.
\subsection{Hierarchical Modality Aggregation and Distribution Network}
The hierarchical modality aggregation and distribution network can be divided into two parts: attention-based shallow feature extraction and feature distribution fusion. Their respective roles are to extract the effective parts of shallow features and effectively fuse deep and shallow features.

Attention-based shallow feature extraction first extracts shallow features from the second layer of the ResNet50~\cite{resnet} backbone network and uses an attention mechanism to fuse the shallow features, which uses CBAM~\cite{cbam} attention module. The module contains a channel attention component and a spatial attention component. The experiments of the CBAM show that places channel attention before spatial attention can achieve better performance. Therefore, we use the same design. To facilitate reader comprehension, we briefly introduce channel attention and spatial attention. For further details, please refer to CBAM~\cite{cbam}.
\begin{figure*}[t]
    \centering
    \includegraphics[width=0.85\textwidth]{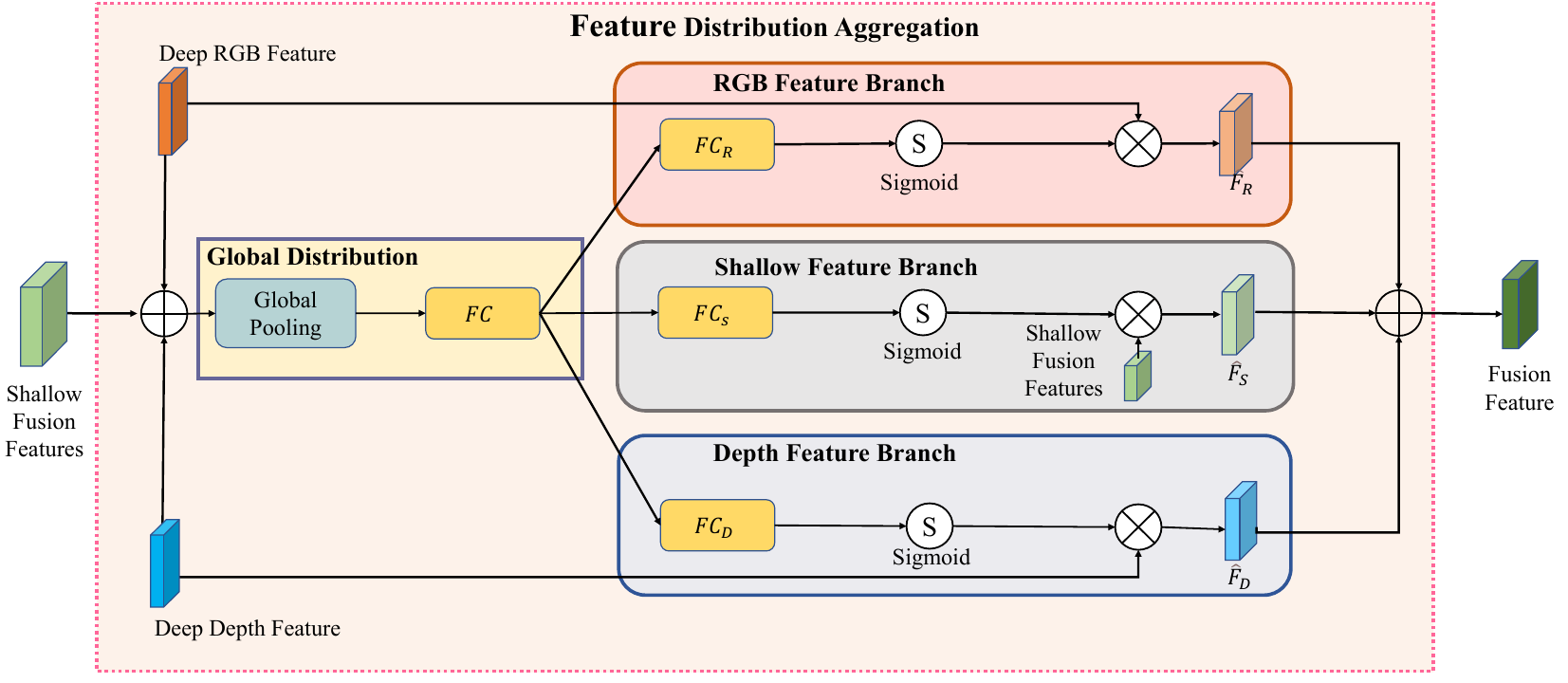}
    \caption{The details of the feature distribution.}
    \label{fig:re}
\end{figure*} 
The channel attention first apply max-pooling and mean-pooling to the features. Subsequently, the channel attention distributes weights across the channels, employing a multi-layer perceptron for this process. The output feature vectors from these two parts are then summed element-wise and activated using a sigmoid function. 
The formulation governing the channel attention mechanism is given as follows:
\begin{equation}
\begin{aligned}
\label{rg_channel}
F_{c} & =\sigma(\operatorname{MLP}(\operatorname{AvgPool}(F_{ori}))+\operatorname{MLP}(\operatorname{MaxPool} (F_{ori}))),
\end{aligned}
\end{equation}
where $F_{c}$ represents the feature processed by the channel attention mechanism, $\sigma(\cdot)$ denotes the sigmoid activation function, and $\operatorname{MLP(\cdot)}$ stands for a multi-layer perceptron; $\operatorname{AvgPool(\cdot)}$ represents average pooling and $\operatorname{MaxPool(\cdot)}$ correspond to max pooling operations. 

The role of spatial attention is to enhance the meaningful local regions within features. The feature output from channel attention is first concatenated after max pooling and average pooling operations in sequence, and then output after convolution and activation operations. 
The spatial attention formulation is as follows:
\begin{equation}
\label{rgspa:1}
F_{s}=\sigma\left(f^{7 \times 7}([\operatorname{AvgPool}(\hat{F}_{c}) ; \operatorname{MaxPool}(\hat{F}_{c})])\right),
\end{equation}
where $F_{s}$ represents the feature processed by the spatial attention mechanism, $\sigma(\cdot)$ denotes the Sigmoid activation function, and $f^{7 \times 7}$ is a convolution operation with a ${7 \times 7}$ kernel. $\operatorname{AvgPool(\cdot)}$ and $\operatorname{MaxPool(\cdot)}$ are the average pooling and max pooling operations respectively.

After the attention mechanism, the parts of shallow features that are beneficial for the tracking task are greatly enhanced. These features not only complement the lack of texture information and detailed features in deep features, but also guide deep features to pay more attention to the parts that require more attention. This module uses convolutional pooling operations to scale it to the same size as the deep features, which are then fed into the feature distribution module together with the deep features.

The inputs to this module include deep RGB features, deep depth features, and shallow fusion features. After adding the three features, they are passed into the global average pooling operation. This operation obtains a global feature vector through average pooling and projection operations to comprehensively consider the three aspects of features. Then this feature vector is divided into three parts, which are the RGB feature branch, the depth feature branch, and the shallow feature branch. Each branch goes through a projection operation and activation operation, and outputs a weight for each channel of the branch. This weight describes in detail the importance of each channel of the three features. Multiplying this weight by the three-channel input feature elementally, the adjusted deep and shallow features can be obtained. These features not only comprehensively consider the necessary feature information globally but also contain contour and texture features, which greatly assist the subsequent tracking tasks.
The formula for global distribution is as follows:
\begin{equation}
\label{rgglo}
F_{g}=FC\left(G P\left(F_R \oplus F_D \oplus F_S\right)\right),
\end{equation}
where $\operatorname{FC(\cdot)}$ represents fully connected layer, $\operatorname{GP(\cdot)}$ represents global pooling; $F_R$, $F_D$ and $F_S$ represent deep RGB features, deep depth features, and shallow fusion features, respectively; $\oplus$ denotes element-wise addition. 
After obtaining the distribution features of the three branches, the final fused features can be obtained again using element-wise addition.

\begin{equation}
\label{rgbranch}
\hat{F}_i=F_i \otimes \sigma\left(FC_i\left(F_{global}\right)\right), i\in\{R, D, S\} ,
\end{equation}
where $\hat{F}_i$ represents the processed corresponding feature; $\operatorname{FC_i(\cdot)}$ represents fully connected layer in different branches; $\otimes$ denotes the element-wise multiplication operation.

\section{EXPERIMENTS}
\begin{table*}
    \centering
    \caption{Comparision between HDMA and the state-of-the-arts trackers. The best results are highlighted in \textcolor{red}{red} and the second are highlighted in \textcolor{green}{green}. The performance is evaluated in terms of precision (Pr), recall (Re), F-score and frame per second (FPS).}
    \label{tbl:table1}
    \begin{tabular}{c|ccc|ccc|c}
        \hline
        \multirow{2}{*}{Method} & \multicolumn{3}{c|}{DepthTrack} & \multicolumn{3}{c|}{RGBD1K} & \multirow{2}{*}{FPS} \\
         & Pr & Re & F-score   & Pr & Re & F-score &\\
        \hline
        Siam\_LTD~\cite{2020vot} & 0.342 & 0.418 & 0.376 & 0.543 & 0.318 & 0.398 & 13.0 \\
        DAL~\cite{dal} & 0.512 & 0.369 & 0.429 & \textcolor{green}{0.562} & 0.407 & 0.472 & - \\
        CLGS\_D~\cite{2020vot}  & 0.369 & 0.584 & 0.453 & - & - & - & 7.3 \\
        ATCAIS~\cite{2020vot} & 0.455 & 0.500 & 0.476 & 0.511 & 0.451 & 0.479 & 1.3 \\
        DDIMP~\cite{2020vot} & 0.469 & 0.503 & 0.485 & 0.557 & 0.534 & 0.545 & 4.7 \\
        Drefine~\cite{2021vot} & - & - & - & 0.532 & 0.462 & 0.494 & - \\
        OSTrack~\cite{ostrack} & 0.522 & 0.536 & 0.529  & - & - & - & - \\
        DeT~\cite{DepthTrack} & 0.506 & 0.560 & 0.532 & 0.438 & 0.419 & 0.428 & \textcolor{green}{36.8} \\
        SPT~\cite{rgbd1k} & 0.549 & 0.527 & 0.538 & 0.545 & \textcolor{red}{0.578} & \textcolor{green}{0.561} & 25.3 \\
        ProTrack~\cite{protrack} & 0.583 & 0.573 & 0.578 & - & - & - & - \\
        ViPT~\cite{VIPT} & \textcolor{green}{0.596} & \textcolor{green}{0.592} & \textcolor{green}{0.594} & - & - & - & 14.1 \\
         \hline
        \textbf{HMAD} & \textcolor{red}{0.626} & \textcolor{red}{0.597} & \textcolor{red}{0.611} & \textcolor{red}{0.573} & \textcolor{green}{0.552} & \textcolor{red}{0.562} & \textcolor{red}{50.0} \\ 
        \hline
    \end{tabular} 
\end{table*}
\subsection{Datasets and Metrics}
We conduct comparison experiments with existing high-performance trackers on two popular RGB-D tracking datasets, DepthTrack~\cite{DepthTrack} and RGBD1K~\cite{rgbd1k}. Given that the datasets predominantly consists of extended temporal sequences, evaluation metrics for long-term tracking have been selected to assess performance. These include precision (Pr), recall (Re), and F-score~\cite{eval1,eval2}.

Precision is calculated by the Gaussian Mixture Distribution between all frame output boxes and the given correct output boxes. The sum of all computed Gaussian Mixture Distributions is divided by the total frame count to determine tracking precision. The precision is calculated as follows:
\begin{equation}
\label{rgequ:1}
\operatorname{Pr}\left(\tau_\theta\right)=\frac{1}{N_p} \sum_{t} \Omega\left(A_t\left(\theta_t\right), G_t\right), t \in\left\{t: A_t\left(\theta_t\right) \neq \emptyset\right\} ,
\end{equation}
where $\operatorname{Pr}\left(\tau_\theta\right)$ 
represents precision, $A_t\left(\theta_t\right)$ represents the tracker's output, $G_t$ represents the ground truth, and 
$\Omega(\cdot)$ represents the intersection of the two. The sum is taken over all non-empty predicted results.

Recall is calculated by the Gaussian Mixture Distribution between all frame output boxes and the given correct output boxes. The sum of all computed Gaussian Mixture Distributions is divided by the total frame count where targets are present to determine tracking recall:
\begin{equation}
\label{rgequ:2}
\operatorname{Re}\left(\tau_\theta\right)=\frac{1}{N_g} \sum_{t} \Omega\left(A_t\left(\theta_t\right), G_t\right), t \in\left\{t: G_t \neq \emptyset\right\} ,
\end{equation}
where $\operatorname{Re}\left(\tau_\theta\right)$ represents recall, $A_t\left(\theta_t\right)$ represents the tracker's output. The sum is taken over all non-empty ground truth results.

F-score is divided by the summary of Pr and Re and then multiplied by two to obtain the tracking F-score:
\begin{equation}
\label{rgequ:3}
F\text{-}score\left(\tau_\theta\right)=2\frac{\operatorname{Pr}\left(\tau_\theta\right) \operatorname{Re}\left(\tau_\theta\right)}{\left(\operatorname{Pr}\left(\tau_\theta\right)+\operatorname{Re}\left(\tau_\theta\right)\right)} ,
\end{equation}
where $\operatorname{Re}\left(\tau_\theta\right)$  represents the corresponding recall;  $\operatorname{Pr}\left(\tau_\theta\right)$ represents the corresponding precision.
\begin{figure}[t]
    \centering
    \includegraphics[width=0.48\textwidth]{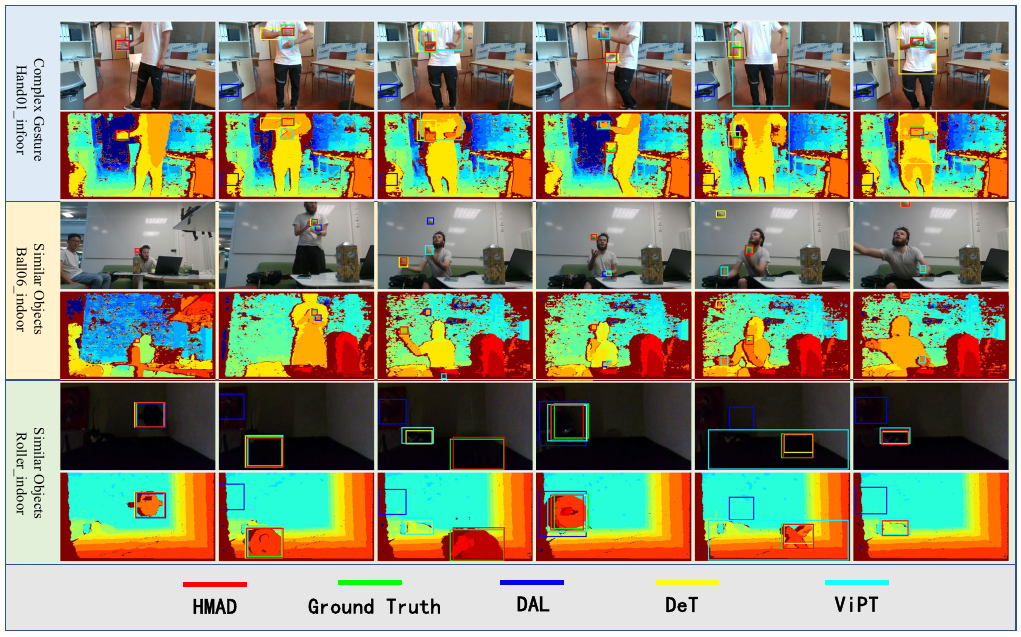}
    \caption{Qualitative comparison between HDMA and other trackers on three challenging sequences in DepthTrack dateset.}
    \label{fig:visual}
\end{figure} 

\subsection{Implementation Details}
All experimental methods are trained on a server with an 5.2GHz CPU and a RTX-3090 GPU with 24GB memory. The proposed tracker was then deployed on the NVIDIA Jetson AGX Orin platform, which boasts a 1.6GHz CPU and 2,000 stream processors GPU. In terms of parameter settings, we utilize an Adam optimizer with a learning rate of 2e-4 for optimization. The network loss function is divided into two parts, namely the classification loss and the IoU loss. Both of these are online update loss functions that change as the model is updated online. During prediction, for the first annotated frame, the initial set of samples is set to 15, the threshold for adding samples is set to 0.6, and the maximum number of samples is set to 50.
\subsection{Compararison with the State-of-the-Arts}
To verify the effectiveness of HMAD, we compare HMAD on the test set of DepthTrack~\cite{DepthTrack} and RGBD1K~\cite{rgbd1k} with 11 state-of-arts methods, including DDIMP~\cite{2020vot}, ATCAIS~\cite{2020vot}, Siam\_LTD~\cite{2020vot}, OSTrack~\cite{ostrack}, CLGS\_D~\cite{2020vot}, Drefine~\cite{2021vot}, DAL~\cite{dal}, DeT~\cite{DepthTrack}, ProTrack~\cite{protrack}, SPT~\cite{rgbd1k} and ViPT~\cite{VIPT}. All the tracking speed are derived from the published papers.

The results of the proposed tracker on DepthTrack~\cite{DepthTrack} are presented in Table~\ref{tbl:table1}. The proposed method outperforms all existing techniques, including the prompt learning based method ViPT~\cite{VIPT}. Specifically, we observe improvements of 3.0\%, 0.5\% and 1.7\% in Pr, Re, and F-score, respectively. Moreover, our method achieves the fastest tracking speed among the existing methods. 

The comparative results on the RGBD1K~\cite{rgbd1k} dataset are shown in Table~\ref{tbl:table1}. The proposed tracker surpasses all other methods. Compared to the second-placed SPT method, our method leads by 2.8\% and 0.1\% in PR and F-score, respectively. The fact that the proposed tracker achieves the best performance on the larger, more complex RGBD1K dataset sufficiently demonstrates its superiority.
\begin{figure}[t]
    \centering
    \includegraphics[scale=0.45]{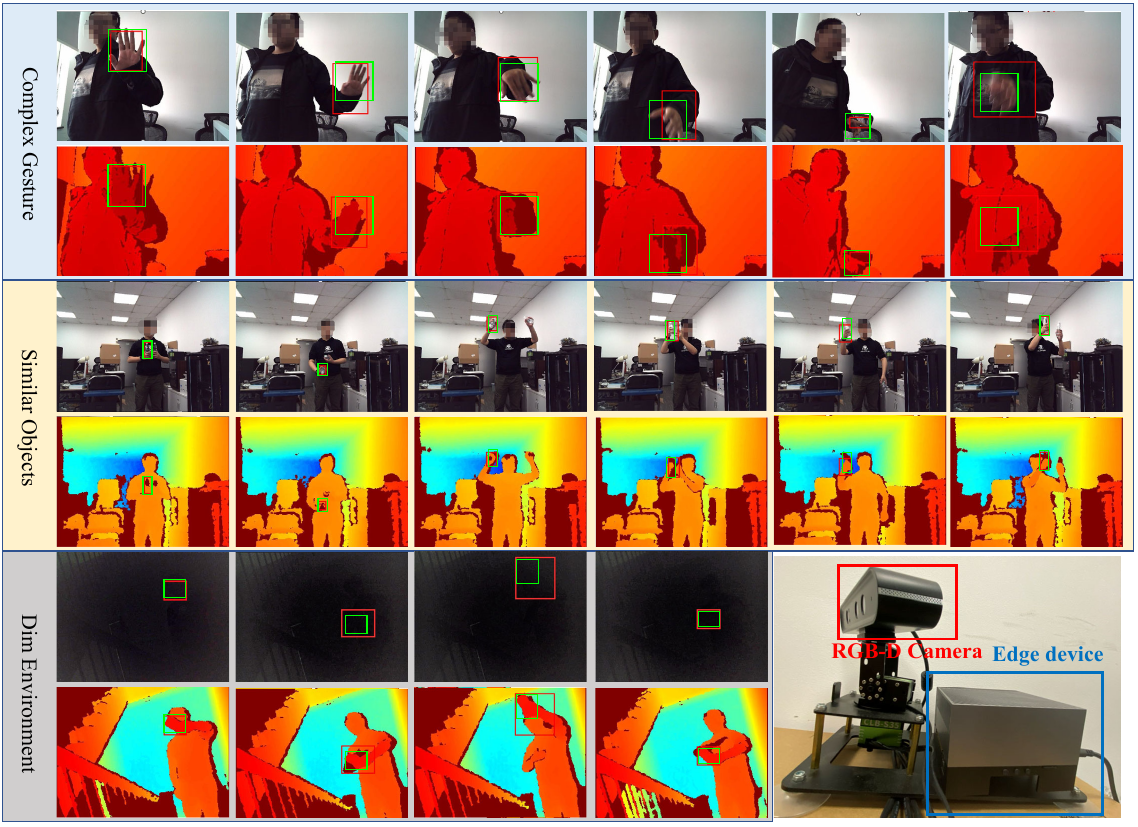}
    \caption{Real-world test results of the proposed tracker, the tracking results are marked in \textcolor{red}{red} and the ground truth are marked in \textcolor{green}{green}.}
    \label{fig:realworld}
\end{figure} 
To intuitively demonstrate the effectiveness of HMAD, we select three challenging sequences from DepthTrack dataset~\cite{DepthTrack} for visualization analysis, as shown in Figure \ref{fig:visual}. In the Hand01\_indoor sequence, the main challenges are complex hand gesture changes and similar targets. HMAD accurately and robustly tracks the target, while other methods suffer from tracking losses and errors. In the Ball06\_indoor sequence, the primary challenges include numerous similar targets and rapid movement, HMAD stably tracked the correct target. In the Rolle\_indoor sequence, the main challenge is the dimly lit environment, and HMAD effectively overcame this problem, achieving stable tracking.
\subsection{Ablation Study}
\begin{table}[t]
\begin{center}
\caption{Ablation study on different components. The performance is evaluated on the DepthTrack test set in terms of precision (Pr), recall (Re) and F-score.} \label{tab:ab}
\begin{tabular}{cccc}
\hline
Variations & Pr & Re & F-score  \\ \hline
baseline& 0.548 & 0.525 & 0.536   \\
\emph{w/o} distribution& 0.571 & 0.545 & 0.557   \\
\emph{w/o} attention& 0.596 & 0.562 & 0.579   \\
HMAD & 0.626 & 0.597 & 0.611   \\ \hline
\end{tabular}
\end{center}
\end{table}
To evaluate the specific impact of each module within our proposed method, we conduct an ablation experiment using the test set of the DepthTrack~\cite{DepthTrack}. Table~\ref{tab:ab} presents the results of our ablation experiment. The baseline model used in this study directly add RGB and depth data, resulting in Pr, Re, and F-score values of 0.548, 0.525 and 0.536 respectively. We use the attention module~\cite{cbam} independently on top of the baseline, resulting in improvements of 2.3\%, 2.0\% and 2.1\% for Precision, Recall, and F-score, respectively. Similarly, the incorporation of the feature distribution aggregation module separately led to performance gains of 4.8\%, 3.7\% and 4.1\% for Precision, Recall, and F-score, respectively. The complete HDMA achieve a score of 0.626, 0.597 and 0.611 respectively. The ablation study allows us to investigate the impact of each module on the overall performance of the proposed method.
\subsection{Real-World Tests}
In this section, we deploy the proposed tracker on an actual edge device to validate its performance in the real world. We conduct experiments using the NVIDIA Jetson AGX Orin platform and the Orbbec Astra PRO RGB-D dual-modality camera, The device used is shown in the bottom right corner of Figure \ref{fig:realworld}. Notably, this camera can effectively detect distances up to eight meters with an error of less than three millimeter, meeting the requirements of real-world application scenarios. Meanwhile, the edge device used in this system has a computing power of approximately one-third of the mainstream GPU RTX-3090, which places higher requirements on real-time performance for the method.

RGB-D tracking is primarily used in the real world for human pose estimation and behavior recognition. To evaluate the performance of the proposed tracker in these application scenarios, we simulated real-world scenarios and added corresponding challenges. As shown in rows one and two of Figure \ref{fig:realworld}, in the commonly used gesture tracking task in human-machine interaction, the proposed method can stably track complexly varying hand targets. The images in row three and four demonstrate tracking in scenes with similar targets; when there are two similar targets in the scene, the proposed method can still stably track the correct target. The last two rows show the tracking effect in a dim environment; even when the RGB modality is essentially ineffective, the proposed method still successfully tracks the target using depth information. 

Throughout all experiments, the tracker consistently maintained a tracking speed of 15 FPS on the edge device. These experiments comprehensively illustrate that the proposed tracker not only attained excellent metrics on offline datasets but also demonstrated real-time tracking capabilities and exceptional performance in real-world environments and on edge device. 
\section{CONCLUSIONS}

In this paper, we proposed a novel HMAD tracker for real-time and robust RGB-D tracking tasks. We proposed a hierarchical modality aggregation and distribution network that efficiently fuses multi-level features from RGB and depth modalities. Experimental results demonstrated that HMAD achieves state-of-the-art performance on multiple datasets. Moreover, real-world test verified that HMAD can effectively handle various challenges presented in real-world and achieves real-time effectiveness.

\begin{acks}
This work was supported by the National Natural Science Foundation of China (62072232), the Key R\&D Project of Jiangsu Province (BE2022138), the Fundamental Research Funds for the Central Universities (021714380026), the Program B for Outstanding Ph.D. candidate of Nanjing University, and the Collaborative Innovation Center of Novel Software Technology and Industrialization.
\end{acks}
\bibliographystyle{ACM-Reference-Format}
\bibliography{sample-base}










\end{document}